\documentclass{article}

\PassOptionsToPackage{numbers,compress}{natbib}

\usepackage[preprint,eandd]{neurips_2026}

\usepackage[utf8]{inputenc} % allow utf-8 input
\usepackage[T1]{fontenc}    % use 8-bit T1 fonts
\usepackage{hyperref}       % hyperlinks
\usepackage{url}            % simple URL typesetting
\usepackage{booktabs}       % professional-quality tables
\usepackage{amsfonts}       % blackboard math symbols
\usepackage{nicefrac}       % compact symbols for 1/2, etc.
\usepackage{microtype}      % microtypography
\usepackage{xcolor}         % colors
\usepackage{textcomp}
\usepackage{graphicx}
\usepackage{multirow}
\usepackage{color}
\usepackage{threeparttable} % for table footnotes
\usepackage{tablefootnote} % for table footnotes
\usepackage{arydshln} % for dashed lines in tables
\usepackage{amssymb} % for checkmark symbol
\usepackage{pifont} % for \ding{55} symbol
\usepackage{wrapfig}
\usepackage{verbatim}
\usepackage{enumitem}
\usepackage{amsmath}
\usepackage{colortbl}
\usepackage{setspace}
\usepackage{hhline}
\usepackage{subcaption}
\usepackage{makecell}
\usepackage{mathrsfs}
\usepackage{tikz}
\usepackage{tabularx}
\usepackage{fontawesome5}
\usepackage{algorithm}
\usepackage{algorithmic}

\definecolor{pkured}{RGB}{140,0,25}
\newcommand{\MYBENCH}{\textsc{SGR-Bench}}  % State-Gated Retrieval benchmark (black in text)
\newcolumntype{C}[1]{>{\centering\arraybackslash}p{#1}}

\title{\MYBENCH: Benchmarking Search Agents on State-Gated Retrieval}

\usepackage{times}
\usepackage{latexsym}
\usepackage[T1]{fontenc}
\usepackage[utf8]{inputenc}
\usepackage{microtype}
\usepackage{inconsolata}
\usepackage{graphicx}
\usepackage{booktabs}
\usepackage{multirow}
\usepackage{amsmath}
\usepackage{xcolor}
\usepackage{enumitem}
\usepackage{subcaption}
\usepackage{pifont}
\usepackage{caption}
\usepackage{url}

\definecolor{bestcol}{HTML}{1a7f37}
\definecolor{secondcol}{HTML}{0969da}
\definecolor{groupblue}{RGB}{235,242,250}
\definecolor{groupgray}{RGB}{242,242,242}
\newcommand{\best}[1]{\textbf{\textcolor{bestcol}{#1}}}

\author{%
  \vspace{-25pt}\\
  \textbf{Ningyuan Li$^{2,*}$\quad Haiyang Shen$^{1,*,\dag}$\quad Mugeng Liu$^{1}$\quad Yudong Han$^{1}$}\\
  \textbf{Zhuofan Shi$^{1}$\quad Sixiong Xie$^{1}$\quad Yun Ma$^{1,\dag}$}\vspace{8pt}\\
  $^{1}$Peking University \\
  $^{2}$Beijing University of Technology \\
  \texttt{\small ningy@emails.bjut.edu.cn, hyshen@stu.pku.edu.cn, mayun@pku.edu.cn} \vspace{8pt}\\
  \vspace{3em}
  $^*$Equal contribution, \textsuperscript{\dag}Corresponding Authors
  \vspace{-48pt} \\
}

\begin{document}

\maketitle

\begin{abstract}
Recent advances in large language models and tool-using agents have expanded the range of benchmarked web tasks. Yet an important class of specialized retrieval tasks remains undercharacterized. On many specialized data-retrieval websites, answer-bearing evidence becomes accessible only after establishing the correct site-specific retrieval state through filters, views, hierarchies, or scopes. We term this capability \emph{state-gated retrieval} (SGR). We introduce \MYBENCH, a benchmark for this setting containing 100 expert-curated tasks spanning six source families and 12 public data ecosystems. Each task requires discovering the appropriate website and configuring its site-specific retrieval state to produce a structured answer. \MYBENCH~ pairs constraint-guided and goal-oriented formulations of the same underlying problems, enabling controlled comparisons between explicit and implicit guidance for state-gated retrieval. We evaluate eight CLI-based agentic LLM systems and three commercial search-agent products. On \MYBENCH, the strongest system reaches only 66.18\% item-level F1, while row-level F1 remains much lower. A manual audit of 156 analyzable failed CLI trajectories shows why: agents often reach a relevant web source, but establish the wrong site-specific retrieval state. Retrieval-scope drift (37.2\%) and criterion mismatch (27.6\%) dominate, whereas final answer composition accounts for only 10.3\%. The dataset and single-case evaluation instructions are available at \url{https://huggingface.co/datasets/PKUAIWeb/SGR-BENCH}.
\end{abstract}

\section{Introduction}

Recent advances in large language models and tool-using agents, together with corresponding advances in benchmark design, have substantially expanded the range of benchmarked web tasks, spanning knowledge-intensive question answering, interactive search, browser-grounded interaction, and multi-step task execution \cite{nakano2022webgptbrowserassistedquestionansweringhuman,yao2023reactsynergizingreasoningacting,schick2023toolformerlanguagemodelsteach,wei2025browsecomp,zhouwebarena,drouin2024workarena,yoran2024assistantbench}. Search agents can now iteratively query the web, access and read documents, and synthesize evidence across multiple sources to address complex information needs. In many professional retrieval settings, however, identifying the appropriate website is only the first step. On specialized data-retrieval websites, answer-bearing evidence is often not exposed on entry pages or under default settings. It becomes accessible only after the correct filters, views, hierarchies, or scopes are set. We refer to this capability as state-gated retrieval (SGR): the ability to identify the appropriate website and configure its site-specific retrieval state so that answer-bearing evidence unavailable under the default state becomes accessible.

State-gated retrieval remains underexplored in existing benchmarks. Search-agent benchmarks such as BrowseComp, WebWalkerQA, and WideSearch mainly evaluate source discovery, search depth and breadth, and cross-page evidence aggregation over the open web \cite{wei2025browsecomp,wu2025webwalker,wong2025widesearch}. DeepSearchQA extends this line toward comprehensive deep research evaluation \cite{gupta2026deepsearchqa}. Web-agent benchmarks such as WebArena, Mind2Web, and WorkArena instead emphasize browser-grounded interaction, action sequencing, and end-to-end task completion \cite{zhouwebarena,deng2023mind2web,drouin2024workarena}. Taken together, these benchmark lines have substantially advanced the evaluation of web-based information seeking. While highly valuable, they still leave open whether an agent can make answer-bearing evidence accessible by establishing the site-specific retrieval state required by a specialized website.

This gap matters especially on specialized data-retrieval websites. Answer-bearing evidence is often governed by the website's site-specific retrieval state rather than exposed through static pages or direct lookup. Agents must therefore map domain constraints to state-setting controls such as filters, views, hierarchies, and result scopes, often under dependencies induced by earlier retrieval steps. In such settings, errors in candidate selection or scope control can easily propagate, causing agents to miss the evidence needed for correct completion.

To close this gap, we introduce \MYBENCH, a benchmark for state-gated retrieval in specialized web retrieval. The current release contains 100 expert-curated tasks spanning six higher-level source families and grounded in 12 public data ecosystems, and is designed to isolate state-gated retrieval in a controlled yet realistic setting. Each task presents a natural-language information need without revealing the target website, requiring the agent to discover the appropriate website, configure its site-specific retrieval state, and produce a structured answer grounded in on-site answer-bearing evidence. \MYBENCH~ pairs constraint-guided and goal-oriented formulations of the same underlying problems, enabling controlled comparisons between explicit and implicit guidance for state-gated retrieval. We further employ rigorous quality control through a systematic data curation pipeline, including candidate website curation, task-design protocol specification, candidate construction and filtering, and multi-round expert validation for answer identifiability, shortcut resistance, and state-gated-retrieval necessity.

Evaluation on \MYBENCH~ reveals a pronounced gap between partial access to answer-bearing evidence and correct structured completion for current search-agent systems. We benchmark eight CLI-based agentic LLM systems, including GPT-5.5 and Claude Opus 4.7 \cite{singh2026openaigpt5card,anthropic2026claudeopus47systemcard}, Gemini 3.1 Pro and Qwen3.6-Plus \cite{googledeepmind2026gemini31proevaluation,yang2025qwen3technicalreport}, GLM-5.1 and Seed-2.0 Pro \cite{glm5team2026glm5vibecodingagentic,bytedanceseed2026seed20modelcard}, and Kimi K2.5 and DeepSeek V4 Pro \cite{kimiteam2026kimik25visualagentic,deepseekai2026deepseekv4technicaldocumentation}, together with three commercial search-agent products: Google Search AI Mode, Gemini Deep Research, and OpenAI Deep Research \cite{google2025aimodesearch,google2026geminideepresearch,openai2025deepresearchsystemcard}. On \MYBENCH, overall Item-F1 ranges from 14.87\% to 66.18\%, while row-level correctness remains substantially lower. Audited CLI failures further show that the dominant errors do not arise from failing to locate a relevant source. Instead, agents often reach the appropriate website but fail to preserve the site-specific retrieval state under which answer-bearing evidence remains valid. Retrieval-scope drift (37.2\%) and criterion mismatch (27.6\%) together account for 64.7\% of audited failures, while final answer composition accounts for only 10.3\%. Taken together, these results indicate that the main bottleneck is not source discovery, but preserving the site-specific retrieval state needed to turn partial access to answer-bearing evidence into correct structured completion.

Our main contributions are threefold:
\begin{itemize}[leftmargin=*,itemsep=1.0pt,topsep=0pt,parsep=0pt,partopsep=0pt]
\item We introduce \MYBENCH, a benchmark explicitly centered on state-gated retrieval in specialized web retrieval. The benchmark contributes 100 expert-curated tasks spanning six higher-level source families and 12 public data ecosystems, with paired constraint-guided and goal-oriented formulations that support diagnosis of explicit and implicit guidance for state-gated retrieval.
\item We establish a systematic data curation methodology for building state-gated-retrieval benchmarks, combining candidate website curation, task-design protocol specification, candidate construction and filtering, and multi-round expert validation under six design requirements: domain specificity, long-tail source grounding, answer uniqueness and verifiability, ground-truth stability, shortcut resistance, and logical dependency.
\item We provide a broad empirical evaluation of current search agents on this setting, covering eight CLI-based agentic LLM systems and three commercial products. The resulting evidence shows that the main failure mode is operating the right website in the wrong site-specific retrieval state, rather than final formatting alone.
\end{itemize}
% ===================================================================

\section{Related Work}

This section reviews prior work from two complementary perspectives. Section~\ref{subsec:search_agent_benchmarks} discusses search-agent benchmarks. Section~\ref{subsec:web_navigation_benchmarks} examines web navigation and interaction benchmarks. Taken together, these perspectives clarify the evaluation gap that \MYBENCH~ is designed to address. Table~\ref{tab:comparison} summarizes the positioning of \MYBENCH~ relative to representative prior benchmarks.

\begin{table}[htbp]
\centering
\caption{Comparison with representative prior benchmarks. \textit{Expert Data-Retrieval Sites}: tasks are grounded in specialized websites. \textit{State-Gated Evidence}: answer-bearing evidence becomes accessible only after the required site-specific retrieval state is established.}
\label{tab:comparison}
\small
\setlength{\tabcolsep}{3.5pt}
\renewcommand{\arraystretch}{1.04}
\begin{tabularx}{\columnwidth}{
l
>{\hsize=0.8\hsize\centering\arraybackslash}X
c
c
>{\hsize=1.2\hsize\centering\arraybackslash}X
}
\toprule
\textbf{Benchmark} &
\textbf{Primary Focus} &
\makecell[c]{\textbf{Expert Data}\\\textbf{Retrieval Sites}} &
\makecell[c]{\textbf{State-Gated}\\\textbf{Evidence}} &
\textbf{Output Target} \\
\midrule
BrowseComp &
Deep search &
\ding{55} &
\ding{55} &
Short answer \\

WebWalkerQA &
Deep search &
\ding{55} &
\ding{55} &
Short answer \\

WideSearch &
Wide search &
\ding{55} &
\ding{55} &
Organized table / list \\

DeepSearchQA &
Deep+wide search &
\ding{55} &
\ding{55} &
Exhaustive answer set \\

WebArena &
Task execution &
\ding{55} &
\ding{55} &
Task completion \\

Mind2Web &
Action grounding &
\ding{55} &
\ding{55} &
Action sequence \\

WorkArena &
\mbox{Enterprise task execution} &
\ding{55} &
\ding{55} &
Task completion \\

\textbf{\MYBENCH} &
\mbox{\textbf{SGR evidence seeking}} &
\ding{51} &
\ding{51} &
\textbf{Exhaustive structured outputs} \\
\bottomrule
\end{tabularx}
\end{table}

\subsection{Search-Agent Benchmarks}
\label{subsec:search_agent_benchmarks}

Search-agent benchmarks increasingly evaluate agent performance on open-web information-seeking tasks. Earlier knowledge-intensive and open-domain question answering benchmarks, such as Natural Questions~\cite{kwiatkowski2019natural}, TriviaQA~\cite{joshi2017triviaqa}, and KILT~\cite{petroni2021kilt}, evaluate retrieval-based answering with evidence drawn from Wikipedia or the open web. More recent search-augmented evaluations, such as FreshQA~\cite{vu2024freshllms} and GAIA~\cite{mialon2023gaia}, move closer to dynamic assistants that rely on web access and tool use. Deep-search benchmarks such as BrowseComp~\cite{wei2025browsecomp} and WebWalkerQA~\cite{wu2025webwalker} focus on search tasks in which answer-bearing evidence is difficult to locate and often requires sustained multi-step browsing over complex web structures. Wide-search benchmarks such as WideSearch~\cite{wong2025widesearch} instead stress broad source coverage and answer-set completeness, requiring agents to exhaustively collect and deduplicate relevant items from large candidate sets. More recent benchmarks such as DeepSearchQA~\cite{gupta2026deepsearchqa} and DeepWideSearch~\cite{lan2025deepwidesearch} combine both dimensions, requiring agents to navigate difficult source structures while assembling exhaustive structured outputs. Concurrent deep-research benchmarks further examine report-level research, rubric-based diagnosis, and cross-domain accuracy in long-horizon web research \cite{du2025deepresearchbenchcomprehensivebenchmark,li2026deepresearchbenchiidiagnosing,zhong2026dracocrossdomainbenchmarkdeep}.

Prior search-agent benchmarks primarily evaluate source discovery, traversal, and cross-page aggregation. \MYBENCH~ extends this line of evaluation by explicitly evaluating a complementary capability: whether an agent can surface answer-bearing evidence that remains inaccessible until the appropriate website is brought into the correct site-specific retrieval state. This capability is important because, on many specialized data-retrieval websites, identifying the appropriate website is necessary but not sufficient, and answer-bearing evidence may remain inaccessible under the default site-specific retrieval state until appropriate state-gated retrieval is applied. Accordingly, \MYBENCH~ adds a distinct evaluation axis to prior search-agent benchmarks by introducing tasks in which agents must identify the appropriate website among expert data-retrieval sources and then access answer-bearing evidence gated behind site-specific retrieval states.

\subsection{Web Navigation and Interaction Benchmarks}
\label{subsec:web_navigation_benchmarks}

Web navigation benchmarks evaluate whether agents can complete tasks effectively in realistic browser environments. Early benchmarks focused on controlled or function-specific web environments. MiniWoB++~\cite{shi2017world} provides synthetic web-interaction tasks, and WebShop~\cite{yao2022webshop} studies goal-directed interaction in an e-commerce setting. Subsequent benchmarks move closer to realistic public-web settings: WebArena~\cite{zhouwebarena} evaluates multi-site task execution in self-hosted web environments, VisualWebArena~\cite{koh2024visualwebarena} introduces visually grounded tasks on realistic websites, and Mind2Web~\cite{deng2023mind2web} uses real interaction traces collected from diverse websites. A further step in this progression emphasizes multi-turn and longer-horizon workflows. WebLINX~\cite{lu2024weblinx} studies multi-turn website navigation from real-world demonstrations, WorkArena~\cite{drouin2024workarena} and WorkArena++~\cite{boisvert2024workarena++} focus on enterprise software and compositional workplace workflows, respectively, and AssistantBench~\cite{yoran2024assistantbench} evaluates realistic and time-consuming web tasks. Recent work has also expanded web-agent evaluation toward deterministic replicas, safety-oriented browser tasks, aligned browser-agent behavior, real-website end-to-end agents such as WebVoyager, and standardized browser-agent ecosystems such as BrowserGym \cite{garg2026real,tur2025safearena,kumar2025aligned,he2024webvoyager,dechezelles2024browsergym}.

Web navigation benchmarks primarily assess browser-grounded task execution, including action grounding and completion of multi-step workflows. Recent benchmarks with realistic websites and replicas evaluate important state-changing workflows, but they do not center expert-curated retrieval tasks whose answer-bearing evidence is hidden behind source-specific data controls. \MYBENCH~ therefore shifts the evaluation target from task execution to information seeking on specialized data-retrieval websites: the relevant question is whether an agent can identify the appropriate website and surface answer-bearing evidence that remains hidden until the site is brought into the correct site-specific retrieval state. Accordingly, our current system comparison targets production search-agent systems rather than purpose-built browser agents, because the benchmark is designed to measure retrieval-state establishment in realistic search workflows rather than general browser-control competence.

% ===================================================================

\section{\MYBENCH}

This section presents \MYBENCH~ from four perspectives. Section~\ref{sec:task-design} defines the state-gated retrieval task and its formal setting. Section~\ref{sec:data-curation} describes the four-stage data curation pipeline. Section~\ref{sec:dataset_statistics} reports dataset statistics and taxonomy. Section~\ref{sec:evaluation} specifies the evaluation protocol and metrics.

\subsection{Task Definition}
\label{sec:task-design}

For each task, let \(W\) denote the target website and let \(s\) denote a site-specific retrieval state of \(W\), determined by website controls such as filters, views, hierarchy selections, and scopes. We write \(V(W,s)\) for the entries, result views, or page content exposed under state \(s\). For a task with answer \(a\), the answer-bearing evidence \(E(a)\) is not exposed under the default state \(s_0\), but becomes accessible along a trajectory of states \(s_1,\ldots,s_k\) induced by state-setting operations. An SGR task therefore requires an agent to identify \(W\) and find an operation sequence such that \(E(a) \subseteq \bigcup_{t=1}^{k} V(W,s_t)\). The core difficulty arises when evidence exposed under one state determines which operation is needed next, forcing the agent to maintain and update the site-specific retrieval state across dependent retrieval steps.

This formulation distinguishes the evaluation focus of \MYBENCH~ from that of existing benchmarks. Search-agent benchmarks primarily evaluate source discovery, traversal, and cross-page aggregation over the open web, whereas web-navigation benchmarks emphasize interface grounding and task execution on websites. SGR instead focuses on state-conditioned evidence exposure within specialized websites: the agent must establish the site-specific retrieval state under which answer-bearing evidence becomes accessible, rather than merely locate relevant pages or execute a predefined sequence of web actions.

\textbf{Task.} \MYBENCH~ evaluates search agents on end-to-end SGR tasks. Given a question \(q\) that specifies an information need and an output schema, but does not reveal the target website, an agent \(\mathcal{M}\) equipped with web search, webpage browsing, and document-access tools must identify \(W\), configure the required site-specific retrieval state, and produce a structured answer \(\hat{a}\) grounded in the answer-bearing evidence exposed along the retrieval trajectory. The model may use any search or browsing strategy, but each reference answer is grounded in the target website, and the intended solution path requires exposing the answer-bearing evidence through the website's site-specific retrieval state.

% \begin{figure*}[t]
% \centering
% \includegraphics[width=\textwidth]{figure1_pipeline.png}
% \caption{Overview of the \MYBENCH~ construction and evaluation pipeline. Candidate target websites are first curated from public high-information-density sources and screened for retrieval-oriented controls. For each retained site, tasks are drafted, filtered, and validated through multi-round expert review to ensure answer identifiability, state-gated-retrieval necessity, and shortcut resistance. During evaluation, the model under test receives only the task specification and target website, and its structured output is scored against the reference answer with exact-match and table-sensitive metrics.}
% \label{fig:pipeline}
% \end{figure*}
\subsection{Data Curation Pipeline}
\label{sec:data-curation}

\begin{figure*}[t]
\centering
\includegraphics[width=\textwidth]{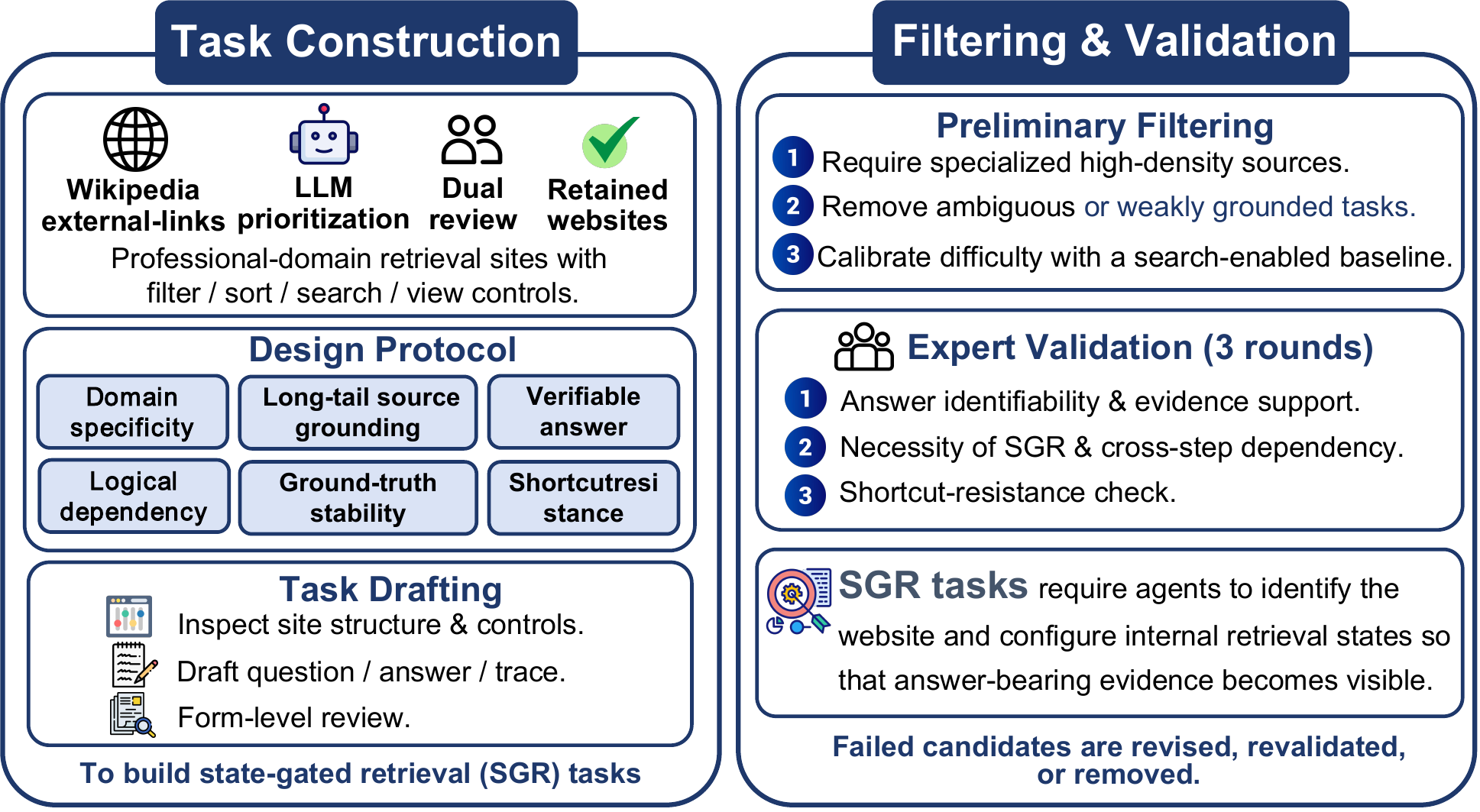}
\caption{Overview of the \MYBENCH~ four-stage data curation pipeline. Candidate websites are drawn from Wikipedia external links, prioritized with an LLM, and retained after dual review. Task candidates are drafted from site structure and retrieval controls under a six-requirement design protocol, then filtered through preliminary screening and three rounds of expert validation for answer identifiability, state-gated-retrieval necessity, and shortcut resistance.}
\label{fig:pipeline}
\end{figure*}

To ensure annotation accuracy while keeping the process cost-effective, we combine LLM-based annotation with human verification in a four-stage pipeline, following the broader use of language models as tool-using assistants in data construction workflows \cite{yao2023reactsynergizingreasoningacting,schick2023toolformerlanguagemodelsteach}. As shown in Figure~\ref{fig:pipeline}, the workflow proceeds through candidate website curation, task design protocol specification, task construction, and candidate filtering and validation. These stages progressively narrow from raw candidate sources to validated benchmark tasks.

\subsubsection{Candidate Website Curation.}
Professional data-retrieval websites are scattered across domains and lack a unified index. 
We therefore use the English Wikipedia external-links dump\footnote{\url{https://dumps.wikimedia.org/enwiki/latest/enwiki-latest-externallinks.sql.gz}} as an initial pool for discovering candidate public data-retrieval websites across domains.
To scale the initial screening, Qwen-Plus prioritizes URLs likely to correspond to information-dense retrieval websites \cite{yang2025qwen3technicalreport}.
Each prioritized candidate then undergoes independent review by two annotators. 
A site is retained only if it supports professional-domain data retrieval rather than merely presenting static text and exposes retrieval-oriented controls such as filtering, sorting, search, or view switching; sites with discrepant judgments are excluded.

\subsubsection{Task Design Protocol}
Before candidate drafting, we define a question-design protocol with six requirements to standardize task construction and support reproducible evaluation.
Concretely, each task must satisfy six requirements: 
\textit{Domain Specificity}, requiring field-specific concepts, terminology, or source knowledge beyond general web search; 
\textit{Long-Tail Source Grounding}, requiring the answer to be located in specialized, high-density retrieval websites rather than common knowledge or direct web lookup; 
\textit{Answer Uniqueness and Verifiability}, requiring a unique answer with unambiguous scoring criteria and source support sufficient for independent verification; 
\textit{Ground-Truth Stability}, requiring the task to be anchored to time-specific or versioned reference sources so that the ground truth remains stable over evaluation; 
\textit{Shortcut Resistance}, requiring the task to avoid directly searchable answers, explicit identifiers that collapse the search space, and bypass paths that eliminate the intended reasoning; 
and \textit{Logical Dependency}, requiring at least one intermediate retrieval result to condition a subsequent search, filtering, or branching decision rather than merely combining independent lookup constraints.

\subsubsection{Task Construction.}
Translating design requirements into concrete tasks requires both domain knowledge about each website's retrieval interface and careful control over question quality. For each curated website, we inspect the site structure and identify retrieval-oriented controls that expose task-relevant site-specific retrieval states. We use ChatGPT-5.2 Pro as a drafting assistant to propose candidate questions, candidate answers, and draft solution traces involving these controls, following recent frontier-model use in agentic task drafting and web research settings \cite{singh2026openaigpt5card,nakano2022webgptbrowserassistedquestionansweringhuman}. We then conduct a form-level review, discarding questions with unnatural phrasing and candidates whose required answer formats are ill-specified or difficult to evaluate reproducibly. Substantive validity is assessed in the subsequent validation stage. Representative task examples are provided in Appendix~\ref{app:examples}.

\subsubsection{Candidate Filtering and Validation.}

\paragraph{Preliminary Filtering.}
We first exclude candidates whose solutions do not materially depend on specialized, high-information-density domain sources, including questions that can be answered without domain-specific terminology or site-specific interpretation.
Within this target scope, we further discard candidates that are ambiguous, underspecified, weakly grounded, otherwise unsuitable for reproducible evaluation, or admit obvious shortcut paths such as direct answer pages or explicit identifiers that collapse the search space.
Finally, we calibrate difficulty using a search-enabled baseline model. We exclude tasks that the baseline solves reliably through near-direct search, as such cases provide limited diagnostic value, and tasks for which the baseline makes no meaningful progress, as these often reflect ill-posed task specifications or unstable answer grounding rather than substantive difficulty. We retain tasks on which the baseline exhibits partial but incomplete progress.

\paragraph{Expert Validation.}
Each candidate that passes preliminary filtering undergoes three rounds of independent expert validation by trained reviewers to assess answer identifiability, state-gated retrieval necessity, and resistance to near-direct shortcut solutions.
In the first round, reviewers solve the task from scratch to verify that the reference answer is factually correct, uniquely identifiable, and supported by answer-bearing source evidence sufficient for independent verification.
In the second round, reviewers verify from the reference solution path that the state-gated retrieval operations are genuinely required for exposing the answer-bearing evidence and induce at least one explicit, nontrivial dependency across retrieval steps.
In the third round, reviewers test shortcut resistance by adversarially probing for residual bypass paths, including pre-existing answer pages, trivial identifiers that collapse the search space, and near-direct external lookup routes that bypass the intended state-gated retrieval operations. This step is designed to reduce the risk that a task can be solved without answer-bearing evidence from the target website or without establishing the required site-specific retrieval state. Candidates that fail any validation round are revised and revalidated or removed.

\subsection{Dataset Statistics}
\label{sec:dataset_statistics}

\paragraph{Scale and source coverage.}
The current release of \MYBENCH~ comprises 100 expert-curated tasks grounded in 12 public data ecosystems and spanning six higher-level source families. At the family level, the distribution is broad but uneven: environmental monitoring accounts for 24 tasks (24.0\%), regulatory resources for 22 (22.0\%), scholarly archives for 18 (18.0\%), life-science resources for 18 (18.0\%), official statistics for 12 (12.0\%), and vulnerability databases for 6 (6.0\%). These sources were selected to cover retrieval interfaces with different state-setting mechanisms, including faceted search, hierarchical browsing, time-window selection, database-specific query fields, and scoped result views.

% \begin{table}[t]
% \centering
% \small
% \begin{tabular}{lc|lc}
% \toprule
% \textbf{Higher-level source family} & \textbf{Count} & \textbf{Higher-level source family} & \textbf{Count} \\
% \midrule
% \makecell[l]{Regulatory\\ resources} & 22 & \makecell[l]{Life-science\\ resources} & 18 \\
% \makecell[l]{Environmental\\ monitoring} & 24 & \makecell[l]{Official\\ statistics} & 12 \\
% \makecell[l]{Scholarly\\ archives} & 18 & \makecell[l]{Vulnerability\\ databases} & 6 \\
% \bottomrule
% \end{tabular}
% \captionsetup{labelformat=empty}
% \caption{Distribution of benchmark tasks across higher-level source families in the current release, counting both prompt variants.}
% \label{tab:data_ecosystems}
% \end{table}

\paragraph{Task taxonomy.}
At the task-taxonomy level, the 100 tasks are evenly split between 50 constraint-guided tasks and 50 goal-oriented tasks. The two variants in each pair are derived from the same information need and are grounded in the same target website, reference answer, evidence requirements, and output format. Constraint-guided variants emphasize the retrieval logic needed to reach the answer, whereas goal-oriented variants emphasize the target information need and leave more of that logic implicit. This paired construction reduces confounding factors and supports cleaner comparisons between explicit and implicit guidance for state-gated retrieval.

\paragraph{Answer schema.}
At the output-schema level, all 100 tasks use ordered-table outputs with prescribed columns and ordering constraints. Reference answers range from 2 to 44 rows, with mean cardinality 6.42 and median 4.0; 72 tasks (72.0\%) require at most seven rows. This unified schema keeps the output space structurally controlled and directly scorable while still spanning a meaningful range of answer-set sizes.

\subsection{Evaluation}
\label{sec:evaluation}

To ensure a fair and consistent comparison across agents, we use reviewer-defined answer canonicalization followed by deterministic metric computation. For each task, trained reviewers specify how raw outputs should be converted into the benchmark schema before scoring. These rules cover concrete cases such as whitespace and punctuation differences, date and unit formatting, capitalization, abbreviation variants, and a small set of task-specific aliases verified against the answer-bearing source evidence. The canonicalization step does not fill in missing fields, correct factual errors, or merge entities that are distinct in the target source.

After canonicalization, each prediction $\hat{a}$ and reference answer $a$ are parsed into structured rows and fields according to the task schema. Rows are aligned by task-specific row keys defined over the primary identifying fields. We report item-level F1, row-level F1, and pairwise order accuracy (P.O.A.). Item-level F1 measures whether individual fields are correct after row alignment, whereas row-level F1 gives credit only when all fields in an aligned row are correct. We additionally report P.O.A. because overlap-based metrics do not capture ordering errors. P.O.A. evaluates whether the relative order among rows shared by the prediction and the reference is preserved, following the pairwise rank-agreement perspective underlying Kendall's $\tau$~\cite{kendall1938new}. Detailed metric definitions and formulas are provided in Appendix~\ref{app:metrics}.

\section{Experiments}
\label{sec:experiments}

\subsection{Experimental Setup}

We evaluate two categories of search-agent systems on \MYBENCH: CLI-based agentic search systems and commercial agent systems. For fair comparison, all agents receive identical prompts specifying the task description, output format constraints, current date, and target query.

\paragraph{CLI-based LLM Agentic Search Systems.}
We evaluate eight frontier LLM-based systems spanning both proprietary and open-weight models, each equipped with search capabilities through a command-line interface (Table~\ref{tab:main_results_two_task_types}). The proprietary models include GPT-5.5 and Claude Opus 4.7 \cite{singh2026openaigpt5card,anthropic2026claudeopus47systemcard}, as well as Gemini 3.1 Pro and Qwen3.6-Plus \cite{googledeepmind2026gemini31proevaluation,yang2025qwen3technicalreport}. The open-weight models include GLM-5.1 and Seed-2.0 Pro \cite{glm5team2026glm5vibecodingagentic,bytedanceseed2026seed20modelcard}, as well as Kimi K2.5 and DeepSeek V4 Pro \cite{kimiteam2026kimik25visualagentic,deepseekai2026deepseekv4technicaldocumentation}. We access GPT-5.5 through the official OpenAI API, while all other models are accessed via the OpenRouter platform \cite{openrouter2026modelsdocs}. For GPT-5.5, we use Codex CLI as the search interface \cite{openai2026codexcli}; for all remaining models, we use Claude Code CLI \cite{anthropic2026claudecodeoverview}. All CLIs run under their default configurations, including medium effort level and thinking mode where configurable. We choose production CLI tools over minimal search-tool wrappers, as default configurations more faithfully reflect the end-user experience and surface issues encountered in practice; results should therefore be interpreted as system-level outcomes rather than model-only rankings.

\paragraph{Commercial Agent Systems.}
In addition to CLI-based implementations, we evaluate closed-source commercial systems to establish a baseline for industrial-grade performance. These systems integrate search, retrieval, and synthesis into end-to-end products behind a unified interface. The evaluated systems are Google Search AI Mode, Gemini Deep Research, and OpenAI Deep Research \cite{google2025aimodesearch,google2026geminideepresearch,openai2025deepresearchsystemcard}. All three are evaluated through manual interaction with their web interfaces. Because each system controls its own retrieval pipeline, we provide only the task prompt and collect the final output.

\paragraph{Metrics.}
We use the evaluation protocol defined in Section~\ref{sec:evaluation}. The main results table reports item-level F1, row-level F1, and pairwise order accuracy (P.O.A.) on \MYBENCH. Overall denotes the average Item-F1 across all 100 tasks.

\subsection{Main Results}
\begin{table*}[t]
\centering
\caption{Main results on \MYBENCH. We report Row-F1, Item-F1, and pairwise order accuracy (P.O.A.); Overall denotes average Item-F1 over all 100 tasks.}
\label{tab:main_results_two_task_types}
\footnotesize
\setlength{\tabcolsep}{2.2pt}
\resizebox{0.82\textwidth}{!}{%
\begin{tabular}{lccccccc}
\toprule
\multirow{3}{*}{\textbf{Model / System}}
& \multicolumn{3}{c}{\textbf{Constraint-Guided Tasks}}
& \multicolumn{3}{c}{\textbf{Goal-Oriented Tasks}}
& \multicolumn{1}{c}{\textbf{Overall}} \\
\cmidrule(lr){2-4} \cmidrule(lr){5-7} \cmidrule(lr){8-8}
& \multicolumn{3}{c}{\textbf{Ordered Table}}
& \multicolumn{3}{c}{\textbf{Ordered Table}}
& \multirow{2}{*}{\textbf{Item-F1}} \\
\cmidrule(lr){2-4} \cmidrule(lr){5-7}
& \textbf{Row-F1}
& \textbf{Item-F1}
& \textbf{P.O.A.}
& \textbf{Row-F1}
& \textbf{Item-F1}
& \textbf{P.O.A.}
\\
\midrule

\midrule
\rowcolor{groupblue}
\multicolumn{8}{c}{\textit{CLI-based LLM Agentic Search Systems}} \\
\midrule
GPT-5.5                     & \best{45.48} & \best{68.22} & \best{90.91} & 41.26 & 64.15 & 89.90 & \best{66.18} \\
Claude Opus 4.7             & 41.52 & 64.35 & 80.81 & 35.50 & 58.41 & 81.82 & 61.38 \\
Gemini 3.1 Pro              & 23.30 & 56.43 & 89.61 & 30.92 & 61.70 & \best{90.91} & 59.06 \\
Qwen3.6-Plus                & 11.11 & 44.25 & 79.29 & 12.12 & 29.50 & 42.42 & 36.88 \\
GLM-5.1                     & 30.54 & 63.17 & 86.87 & 36.09 & \best{68.10} & 87.63 & 65.64 \\
Seed-2.0 Pro                & 22.73 & 31.48 & 45.45 & 18.18 & 28.27 & 27.27 & 29.88 \\
Kimi K2.5                   & 30.19 & 47.54 & 81.82 & 37.28 & 45.17	& 72.73 & 47.39 \\
DeepSeek V4 Pro             & 25.50 & 56.67 & 80.00 & 39.78 & 65.29 & 90.00 & 60.98 \\
\midrule
\rowcolor{groupblue}
\multicolumn{8}{c}{\textit{Commercial Agent Systems}} \\
\midrule
Gemini Deep Research        & 11.11 & 29.72 & 45.45 & 9.42	& 30.15	& 45.45 & 29.93 \\
Google Search AI Mode       & 1.40 & 16.73 & 51.52 & 3.31 & 13.01 & 36.36 & 14.87 \\
OpenAI Deep Research        & 39.67 & 57.27 & 71.72 & \best{43.33} & 51.14 & 62.63 & 54.20 \\

\bottomrule
\end{tabular}
}
\end{table*}

Table~\ref{tab:main_results_two_task_types} summarizes the main results. We focus on three questions tied to web retrieval: whether agents can keep retrieved rows bound to the site-specific retrieval state under which they were obtained, where failures first enter the within-site retrieval process, and which properties of web data interfaces make SGR difficult.

\begin{figure*}[t]
\centering
\includegraphics[width=\textwidth]{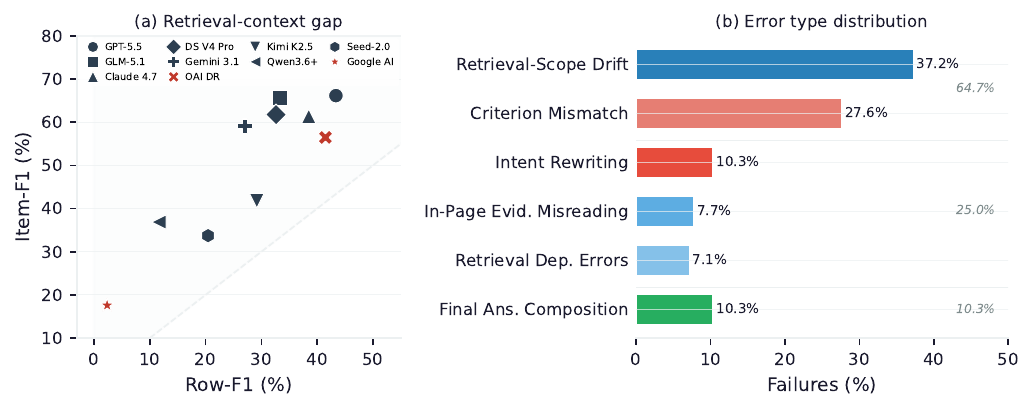}
\caption{(a) Item-F1 vs. Row-F1 for all systems. All points sit above the diagonal: agents recover field values more often than complete rows. (b) Error type distribution across 156 analyzable failed CLI trajectories. Most failures enter while configuring or maintaining the website's site-specific retrieval state (64.7\%), rather than during final answer composition (10.3\%).}
\label{fig:combined}
\end{figure*}

\paragraph{Finding 1: Answer-bearing evidence is found, but the site-specific retrieval state is not preserved.}
Across all evaluated systems, Item-F1 ranges from 14.87\% to 66.18\% (mean 47.85\%), showing that agents often reach pages or result entries containing some of the required field values. Row-F1 is much lower, with a mean of 26.81\%, yielding a 21.04-point gap. Even the strongest system has a 22.81-point gap (66.18\% Item-F1 vs. 43.37\% Row-F1). The same separation appears in both constraint-guided tasks (48.71\% Item-F1 vs. 25.69\% Row-F1) and goal-oriented tasks (46.81\% vs. 27.93\%). This pattern is consistent with retrieval-state loss: agents can copy locally correct values from a page, but the trajectory audit below shows that failures usually enter when they do not preserve the site-specific retrieval state that made those values valid, including the active filters, selected hierarchy node, result scope, and row identity. Figure~\ref{fig:combined}(a) visualizes this gap.

\paragraph{Finding 2: The hard step is configuring the website, not finding the website.}
We manually audited 176 trace-bearing trajectories from eight CLI-based agents and analyzed 156 failed runs. The dominant failure modes are \textit{Retrieval-Scope Drift} (37.2\%) and \textit{Criterion Mismatch} (27.6\%), together accounting for 64.7\% of audited failures. This identifies within-site state control, rather than final answer assembly, as the central bottleneck (Figure~\ref{fig:combined}(b); Appendix~\ref{app:error_defs}).

\paragraph{Finding 3: Hard sites require keeping several web controls aligned.}
\begin{wrapfigure}{r}{0.44\columnwidth}
\centering
\includegraphics[width=0.44\columnwidth]{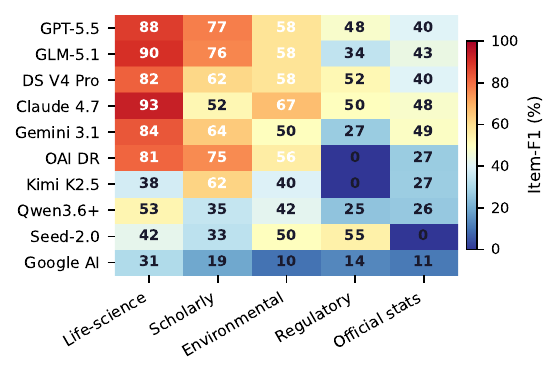}
\caption{Item-F1 (\%) by model and source family on the 100-task benchmark.}
\label{fig:family_heatmap}
\end{wrapfigure}
Figure~\ref{fig:family_heatmap} provides a source-family breakdown of Item-F1 on the 100-task benchmark. Scholarly archives are most accessible (63.1\%), followed by life-science resources (57.5\%) and environmental monitoring (52.9\%). Regulatory resources and official statistics are markedly harder (36.4\% and 34.3\%). The hard cases share a concrete web-retrieval pattern: the answer is valid only when several controls stay aligned--such as agency, jurisdiction, reporting period, population, product class, table universe, or download scope. Once one control drifts, agents still retrieve plausible evidence, but from the wrong slice of the website. This matches the trajectory audit and explains why field-level evidence can look credible while rows are out of scope.

Error-type analysis by source family further explains why these difficulty differences arise (Appendix~\ref{app:error_family}): scholarly and environmental tasks are drift-heavy, whereas regulatory and official-statistics tasks are criterion-mismatch-heavy.

CLI-based systems also outperform commercial systems on mean Item-F1 (53.42\% vs.\ 33.00\%), indicating that stronger interaction control still matters in this setting.
Constraint-guided tasks also yield slightly higher Item-F1 ($\Delta=1.90$), while goal-oriented tasks achieve marginally higher Row-F1 ($\Delta=2.24$), suggesting that explicit guidance helps locate fields but does not solve scope maintenance.

\section{Conclusion}

\MYBENCH~ introduces state-gated retrieval as a benchmark target for specialized web retrieval. Across 100 tasks spanning six source families and 12 public data ecosystems, the best system reaches 66.18\% Item-F1 but much lower Row-F1, showing that partial evidence access often fails to yield correct structured completion. Trajectory audits show that the main bottleneck is within-site state control rather than source discovery. More broadly, the benchmark suggests that progress on specialized web retrieval will require agents to preserve active filters, scopes, and row identities across dependent retrieval steps. It also motivates training and evaluation setups that stress retrieval-state preservation rather than final-answer plausibility alone. This challenge is especially visible on interfaces where multiple filters, jurisdictions, or result scopes must remain aligned throughout the retrieval process. In other words, locally plausible evidence remains insufficient unless agents preserve the exact retrieval context that makes each extracted row valid. The wide gap between scholarly archives and regulatory or official-statistics sites also shows that benchmark difficulty is not driven by source obscurity alone: performance drops most when agents must preserve interacting controls over jurisdiction, reporting period, population, and table scope. This makes SGR a concrete target for interface-aware retrieval research rather than stronger answer synthesis alone. It also suggests that future training data should couple navigation decisions, active filters, and structured extraction into a single supervision signal. This also motivates evaluation protocols that verify whether retrieved rows remain anchored to the correct site slice, since answer-only scoring can mask scope drift behind locally correct fields. 

\noindent\textbf{Limitations.} The current release focuses on public, relatively stable, structured sources and therefore underrepresents highly dynamic web content. Fine-grained error analysis is limited to trace-bearing CLI systems, and we do not yet provide unified trajectory-level scoring across heterogeneous agents. The benchmark is intended for controlled evaluation rather than as a standalone resource for large-scale post-training; Appendix~\ref{app:limitations} gives a fuller discussion.

% \section*{Ethics Statement}

% All benchmark tasks are grounded in publicly accessible data websites, and no private, personally identifiable, or proprietary information is included in the dataset. Task construction involved student participants from Peking University who were informed about the research purpose and consented to contribute. All reference answers are independently verifiable through the cited public sources. The benchmark evaluates model capabilities and does not involve human subjects as evaluation targets. We release the dataset under a permissive license to support reproducible research.
% ===================================================================

\bibliography{references}
\bibliographystyle{plain}

\appendix

\section{Metric Definitions}
\label{app:metrics}

All metrics are computed after reviewer-defined answer canonicalization and task-specific row alignment. Let the canonical reference answer be represented as a structured row sequence $\mathcal{R}_{\mathrm{gold}} = (r_1, \ldots, r_m)$ and the canonical prediction as $\mathcal{R}_{\mathrm{pred}} = (\hat{r}_1, \ldots, \hat{r}_n)$, where each row is a tuple of field values under the task schema. Rows are aligned by task-specific row keys defined over the primary identifying fields, and all field-level comparisons are performed after this alignment, following the structured-output evaluation style used in recent broad and deep web-search benchmarks \cite{wong2025widesearch,gupta2026deepsearchqa,li2026deepresearchbenchiidiagnosing}.

\paragraph{Item-level F1.}
Let $C_{\mathrm{item}}$ denote the number of semantically correct field slots after row alignment, and let $N_{\mathrm{gold}}$ and $N_{\mathrm{pred}}$ denote the total numbers of field slots in the reference and prediction, respectively. We define
\[
\mathrm{Item\mbox{-}F1} = \frac{2 C_{\mathrm{item}}}{N_{\mathrm{gold}} + N_{\mathrm{pred}}}.
\]
This metric gives partial credit when some but not all field values are correct.

\paragraph{Row-level F1.}
Let $C_{\mathrm{row}}$ denote the number of aligned rows whose fields are all semantically correct. We define
\[
\mathrm{Row\mbox{-}F1} = \frac{2 C_{\mathrm{row}}}{|\mathcal{R}_{\mathrm{gold}}| + |\mathcal{R}_{\mathrm{pred}}|}.
\]
A row contributes to $C_{\mathrm{row}}$ only if all of its fields match the reference after normalization and alignment.

\paragraph{Pairwise Order Accuracy (P.O.A.).}
Let $\mathcal{S}$ denote the set of row keys shared by the prediction and the reference. Let $\pi(u)$ and $\hat{\pi}(u)$ denote the positions of a shared row key $u \in \mathcal{S}$ in the reference and prediction, respectively, and let
\[
\mathcal{P}(\mathcal{S}) = \{(u,v) : u,v \in \mathcal{S},\ \pi(u) < \pi(v)\}
\]
be the set of comparable ordered row pairs induced by the reference order. We define
\[
\mathrm{P.O.A.} =
\begin{cases}
\frac{1}{|\mathcal{P}(\mathcal{S})|}
\sum_{(u,v)\in\mathcal{P}(\mathcal{S})}
\mathbf{1}\!\left[\hat{\pi}(u) < \hat{\pi}(v)\right],
& |\mathcal{S}| \ge 2, \\[6pt]
0,
& |\mathcal{S}| < 2.
\end{cases}
\]
P.O.A. therefore measures the fraction of shared-row pairs whose relative order is preserved by the prediction, following the pairwise rank-agreement perspective underlying Kendall's $\tau$ \cite{kendall1938new}. When fewer than two row keys are shared, relative order is not evaluable; in this case, we assign $\mathrm{P.O.A.}=0$ as a conservative convention, treating the prediction as providing no correct ordering information.

\section{Task Examples}
\label{app:examples}

We illustrate the paired task design with an example from the Europe PMC ecosystem. Both variants share the same reference answer, evidence requirements, and output schema; they differ only in how much retrieval procedure is made explicit in the prompt.

\paragraph{Constraint-guided variant (europemc\_003).}
The instruction decomposes the retrieval into explicit steps: (1) find the 2023 and 2024 open-access \textit{Antibodies to Watch} articles on Europe PMC; (2) extract all candidates predicted to file first marketing applications in 2022--2023 from the 2023 article; (3) cross-reference each candidate against the 2024 article and retain only those that have received first approval or entered first regulatory review; (4) output each qualifying molecule with its 2023 category, 2023 prediction label, 2024 status (A for approved, R for under review), 2024 indication, and 2024 region, sorted alphabetically by INN.

\paragraph{Goal-oriented variant (europemc\_003-g).}
The instruction states the same information need but omits the step-by-step decomposition: ``I want to see which candidate molecules predicted in the 2023 \textit{Antibodies to Watch} to file in 2022--2023 have actually been approved or entered first regulatory review in the 2024 article of the same series.'' The agent must independently determine how to locate both articles, extract the 2023 forecast cohort, and reconcile it against the 2024 outcomes.

\paragraph{Reference answer.}
The reference answer contains 8 rows, each representing one qualifying antibody candidate. The answer requires cross-referencing two separate article tables: Table~1 (first approvals) and Table~2 (first regulatory reviews) from the 2024 article, matched against the 2023 forecast baseline. This inter-table dependency is a representative instance of the logical-dependency design requirement.

\section{Detailed Experimental Setup}
\label{app:setup}

\paragraph{Controlled CLI evaluation scope.}
The configuration described here applies to the eight CLI-based systems used in the main evaluation: Kimi K2.5, GLM-5.1, Qwen3.6-Plus, DeepSeek V4 Pro, Seed-2.0 Pro, Claude Opus 4.7, Gemini 3.1 Pro, and GPT-5.5. All eight are evaluated under a common search--fetch--PDF retrieval setup. Observed performance differences therefore reflect system behavior under a matched retrieval-tool regime rather than differences in the available external tools.

\paragraph{Prompt template and controlled protocol.}
All agents receive the same task prompt, consisting of the current date, the benchmark task instruction, and the required structured output schema. Prompts do not provide a start URL, and they instruct agents to solve the task through the designated search, fetch, and PDF tools while prohibiting alternative retrieval paths.

\paragraph{Exposed tools and runtime restrictions.}
Across all eight CLI systems, the runtime exposes only three external retrieval tool classes: Serper search, fetch-based webpage reading, and PDF reading. For the Claude Code systems, each run uses a project-level MCP configuration under a strict MCP configuration, so that only the designated search, fetch, and PDF tools are available during execution. A companion settings file additionally disables the Claude-native \texttt{WebSearch} and \texttt{WebFetch} tools, ensuring that all evaluated systems operate under the same external retrieval-tool budget. GPT-5.5 is executed through Codex CLI rather than Claude Code CLI, but its implementation is aligned to the same evaluation controls, including prompt constraints, exposed retrieval tools, task isolation, and runtime budget.

\paragraph{Execution budget and task isolation.}
Each task is executed in an independent session, with no context shared across tasks. The maximum runtime per task is 6000 seconds; runs are marked as stalled after 3000 seconds without trace progress; and answer extraction is polled every 5 seconds during execution. All CLI-based systems are run at medium effort.

\paragraph{Commercial system evaluation.}
Google Search AI Mode, Gemini Deep Research, and OpenAI Deep Research \cite{google2025aimodesearch,google2026geminideepresearch,openai2025deepresearchsystemcard} are evaluated through manual interaction with their respective web interfaces. The task prompt is provided as-is, and the final output is collected without modification. No intermediate trajectory data is available for these systems.

\paragraph{Public release and reproduction instructions.}
The public release separates benchmark data from evaluation infrastructure. The Hugging Face dataset contains the English benchmark tasks and reference answers used for evaluation, together with documentation for loading the data. Companion reproduction instructions describe the MCP configuration, required API keys, task loading procedure, single-case execution protocol, and single-case scoring procedure. We do not release the full internal batch-running infrastructure or the complete trajectory corpus, since those artifacts include implementation-specific orchestration details and mixed-language traces that are not part of the benchmark definition. Instead, the release focuses on the dataset itself and on a minimal, inspectable path for reproducing individual benchmark cases.

\section{Extended Limitations Discussion}
\label{app:limitations}

The current release of \MYBENCH~ reflects deliberate trade-offs among coverage, diagnostic precision, and reproducibility. These choices make the benchmark suitable for controlled evaluation of state-gated retrieval, but they also define its present scope and leave several important settings underexplored.

\paragraph{Coverage favors stable public interfaces over rapidly changing web content.}
To ensure stable ground truth, reproducible evidence verification, and reliable scoring, we prioritize public sources whose relevant tables, records, and filtering logic remain comparatively stable over time. This design choice improves annotation quality and benchmark longevity, but it underrepresents retrieval settings driven by highly dynamic content, such as breaking news, live operational dashboards, rapidly refreshed public records, or interfaces whose ranking and availability change substantially within short time windows. As a result, \MYBENCH~ should be interpreted as a benchmark for \emph{structured, stateful retrieval under relatively stable public interfaces}, rather than as a comprehensive proxy for all real-time web-search environments.

\paragraph{The evaluation protocol emphasizes final structured outputs rather than unified trajectory-level scoring.}
Our main evaluation centers on the correctness of the final structured answer. This choice aligns with the benchmark's core objective and enables model-agnostic comparison across heterogeneous systems. We complement these outcome metrics with manual trajectory audits, which help localize dominant failure modes such as retrieval-scope drift and criterion mismatch. However, the current release does not yet provide a unified trajectory-level evaluation framework that can systematically score intermediate behaviors across systems, including query reformulation, page selection, filter manipulation, branching decisions, or retrieval-state transitions. Developing such a framework remains an important direction for future work, especially for studying how errors emerge before they become visible in final outputs.

\paragraph{Benchmark scale is intentionally controlled, and is not designed for large-scale post-training by itself.}
Because benchmark construction requires domain-specific candidate discovery, task-specific decomposition, reference-answer verification, and expert review of shortcut resistance, the current release remains modest in scale relative to corpora intended for large-scale model optimization. This scale is sufficient for controlled benchmarking and comparative diagnosis, which are the primary goals of \MYBENCH. At the same time, it means that the benchmark is not intended to serve as a standalone resource for high-volume post-training pipelines, including reinforcement-learning-based optimization, which typically require substantially larger and more diverse task collections. In that sense, \MYBENCH~ is better viewed as an evaluation and diagnosis resource than as a self-sufficient training corpus.

\paragraph{LLM-assisted drafting may still introduce construction-time biases.}
LLMs are used only in a restricted supporting role during benchmark construction, such as candidate prioritization and draft generation, while human experts retain responsibility for website verification, answer validation, shortcut-resistance checks, and final inclusion decisions. Even under this workflow, the benchmark may still inherit subtle drafting biases from the assisting models, including biases in website selection, question phrasing, candidate structuring, or initial decomposition style. Human review substantially reduces these risks and serves as the primary quality-control mechanism, but it cannot guarantee complete removal of all construction-time biases. This limitation should be kept in mind when interpreting both source-family coverage and the stylistic regularities of benchmark tasks.

\section{Full Per-Model Results}
\label{app:domain}

Table~\ref{tab:full_model_results} reports per-model averages across all tasks for the evaluated CLI-based and commercial systems.

\begin{table}[h]
\centering
\small
\caption{Per-model average performance on all tasks, sorted by overall Item-F1.}
\label{tab:full_model_results}
\begin{tabular}{lccc}
\toprule
\textbf{Model} & \textbf{Item-F1} & \textbf{Row-F1} & \textbf{P.O.A.} \\
\midrule
GPT-5.5              & 66.18 & 43.37 & 90.40 \\
GLM-5.1              & 65.64 & 33.32 & 87.25 \\
Claude Opus 4.7      & 61.38 & 38.51 & 81.31 \\
DeepSeek V4 Pro      & 60.98 & 32.64 & 85.00 \\
Gemini 3.1 Pro       & 59.06 & 27.11 & 90.26 \\
OpenAI Deep Research  & 54.20 & 41.50 & 67.17 \\
Kimi K2.5            & 47.39 & 33.74 & 77.28 \\
Qwen3.6-Plus         & 36.88 & 11.62 & 60.86 \\
Gemini Deep Research & 29.93 & 10.27 & 45.45 \\
Seed-2.0 Pro         & 29.88 & 20.45 & 36.36 \\
Google Search AI Mode & 14.87 &  2.36 & 43.94 \\
\bottomrule
\end{tabular}
\end{table}

The results reveal a clear tier structure. The top tier (GPT-5.5, GLM-5.1, Claude Opus 4.7, DeepSeek V4 Pro) achieves Item-F1 above 60\%, with GPT-5.5 leading on both Item-F1 and Row-F1. The middle tier (Gemini 3.1 Pro, OpenAI Deep Research) clusters around 54--59\% Item-F1. The bottom tier (Kimi K2.5, Qwen3.6-Plus, Gemini Deep Research, Seed-2.0 Pro, Google Search AI Mode) falls below 48\%, with Google Search AI Mode achieving only 14.87\% Item-F1.

Notably, OpenAI Deep Research achieves the second-highest Row-F1 (41.50\%) despite ranking sixth in Item-F1 (54.20\%), suggesting that commercial deep-research systems may be better at assembling complete rows from the evidence they do retrieve. Conversely, Gemini 3.1 Pro achieves high P.O.A. (90.26\%) but relatively low Row-F1 (27.11\%), indicating correct ordering of retrieved rows but frequent row-level omissions.

\section{Extended Case Studies}
\label{app:cases}

We present six case studies: one success case, one failure case from the main text, and four additional cases illustrating each major error type.

\paragraph{Success case: GPT-5.5 on reptile\_001.}
This task requires cross-referencing Boulenger's 1890 Indian snake descriptions against the Reptile Database to determine current nomenclature and type specimen status for each species. The reference answer contains 8 species with original names, current names, type status (UNIQUE or AMBIG), and British Museum specimen numbers.

GPT-5.5 achieved perfect scores (Item-F1 = Row-F1 = P.O.A. = 100\%). The trajectory shows the agent correctly: (1) searched for ``Boulenger 1890 India snakes'' on the Reptile Database; (2) configured the advanced search with Author = Boulenger, Year = 1890, Distribution = India; (3) visited each species page to extract current nomenclature and type information; (4) classified each as UNIQUE or AMBIG based on type-specimen information. This trajectory demonstrates successful state-gated retrieval: the agent identified the correct website, configured the site-specific retrieval state through conjunctive filter constraints, and maintained consistent scope across all 8 species pages.

\paragraph{Failure case: GPT-5.5 on waterquality\_003-g.}
This goal-oriented task requires retrieving water quality monitoring data from the USGS Water Quality Portal for specific HUC codes and parameters. GPT-5.5 scored zero on all metrics, with no common row identifiers between its output and the reference.

Trajectory analysis reveals a retrieval-scope drift failure: the agent identified the correct website (Water Quality Portal) but configured incorrect filter parameters, retrieving data for wrong monitoring stations. This error propagated through the entire output, making all subsequent extraction and formatting irrelevant. The case illustrates how a single upstream state-configuration error can invalidate an otherwise well-structured retrieval trajectory.

\paragraph{Retrieval-scope drift: Seed-2.0 on reptile\_001.}
This task required querying the Reptile Database using a structured advanced search (Author=Boulenger, Year=1890, Distribution=India) to enumerate all 12 candidate species and extract taxonomic status fields from each official species page. The agent bypassed the required official root query entirely and instead relied on general-purpose search engine snippets to identify candidates, reducing the 12-candidate systematic pipeline to a single-object guessing workflow. This scope collapse propagated through every downstream stage: only 2 of 12 species pages were visited, the sole output row contained incorrect synonym and type-specimen fields, and 7 of 8 gold rows were entirely absent (Item-F1 = 11.1\%, Row-F1 = 0\%).

\paragraph{Criterion mismatch: Gemini 3.1 on consumerfinance\_012.}
This task required querying the CFPB Complaint Database with the filter \texttt{search\_term = medical OR doctor} to obtain a 585-complaint cohort, then performing hierarchical aggregation to identify peak complaint months, top states, and top companies. The earliest error occurred when the agent translated the textual search criterion into the API query \texttt{medical bills OR doctors} instead of the required \texttt{medical OR doctor}, yielding a spurious 2,311-complaint cohort. Although the agent successfully completed the full analytical pipeline, every downstream computation operated on the incorrect base population. All output slots diverged from the reference (Item-F1 = 0\%, Row-F1 = 0\%).

\paragraph{Intent rewriting: Claude 4.7 on arxiv\_003.}
This task required an exhaustive search of arXiv for papers with v1 submissions in 2020~Q4 under cs.CV whose titles or abstracts mention Vision Transformer variants, followed by per-paper verification of submission history and publication trail. The primary model delegated the entire task to a sub-agent with a prompt that replaced the required exhaustive search with a pre-seeded list of ``well-known early Vision Transformer papers,'' omitting two gold-standard papers (2101.01097 and 2011.08019) from the candidate space entirely. The sub-agent returned a 3-row summary that the primary model accepted without secondary verification, permanently fixing the incomplete candidate set as the final answer (Item-F1 = 68.8\%, Row-F1 = 50.0\%).

\paragraph{Retrieval dependency: Kimi K2.5 on europemc\_002.}
This task required cross-referencing three annual Alzheimer's drug pipeline reviews (2021--2023) from Europe PMC, tracking 24 Phase~1 clinical trials by NCT identifiers to detect phase transitions and drug name changes. The agent successfully located all three correct PMC articles and extracted the relevant tables, including the critical evidence that NCT03634007 (originally AAVrh.10hAPOE2, renamed LX1001) advanced to Phase~2 in 2023. However, during local evidence integration, the analysis script failed to map the 2022 entry back to the 2021 baseline due to the drug name change, erroneously removing it from the candidate set. The final answer contained 5 of 6 gold rows but missed the single most diagnostically important trial, the only one that changed phase across years (Row-F1 = 90.9\%).

\section{Trajectory Audit Categories}
\label{app:error_defs}

To localize where failures arise, we manually audited 176 trace-bearing trajectories from eight CLI-based agents over 22 randomly sampled aligned task slots. We focus on CLI systems because they expose complete, comparable traces, whereas commercial products do not provide sufficient intermediate trajectory data for the same analysis. We use this audit to characterize recurrent failure patterns in the audited CLI subset.

For the audited CLI trajectories, each failed run is assigned one primary error label corresponding to the earliest non-recoverable root cause, with upstream failures taking precedence over downstream answer-writing errors. We use the following six categories.

\paragraph{Retrieval-Scope Drift.}
The agent fails to establish or preserve the correct retrieval workspace, such as the required object set, candidate space, jurisdiction, time range, or result scope. Under this label, later extraction steps may be locally plausible, but they are grounded in the wrong slice of the source.

\paragraph{Criterion Mismatch.}
The agent reaches relevant resources but applies an incorrect decision rule, field definition, aggregation level, denominator, phase space, or time window. The resulting answers often contain locally correct values, but they are bound to the wrong evaluative criterion.

\paragraph{Intent Rewriting.}
The agent implicitly replaces the original task with an easier but non-equivalent surrogate objective, typically by dropping required constraints, weakening exhaustive search requirements, or converting structured retrieval into approximate summarization.

\paragraph{In-Page Evidence Misreading.}
The agent reaches the correct page or table but misreads a local value, label, priority rule, or field interpretation on that page. This category is reserved for errors that can be localized to incorrect page-internal evidence reading rather than to broader scope or criterion selection.

\paragraph{Retrieval Dependency Errors.}
The agent obtains key intermediate evidence but fails to close the dependency chain needed for correct completion, such as cross-page alignment, backtracking, evidence propagation, or final candidate arbitration. The central issue is not missing access, but incomplete dependency resolution across retrieval steps.

\paragraph{Final Answer Composition.}
The upstream retrieval and interpretation are largely correct, but the final answer is corrupted during aggregation, slot mapping, sorting, normalization, deduplication, or transcription. We assign this label only when earlier stages are substantially correct and the failure is concentrated in the final answer assembly step.

\section{Error Profile by Model}
\label{app:error_model}

\begin{figure}[h]
\centering
\includegraphics[width=\columnwidth]{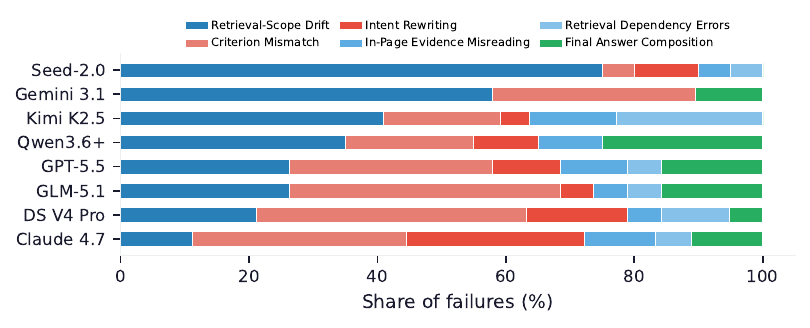}
\caption{Share of each error type within each model's failures. Models are sorted by retrieval-scope drift share (descending). Seed-2.0 and Gemini~3.1 are dominated by scope drift; DS~V4~Pro and GLM-5.1 concentrate on criterion mismatch; Claude~4.7 uniquely concentrates on intent rewriting.}
\label{fig:error_by_model}
\end{figure}

Table~\ref{tab:error_by_model} and Figure~\ref{fig:error_by_model} decompose the six error types across the eight CLI-based models with trajectory data. The 176 annotated trajectories include 156 failures and 20 correct cases.

\begin{table}[h]
\centering
\small
\caption{Error type counts per model across 176 annotated trajectories (22 task slots per model). Percentages are computed over failures only (excluding Correct).}
\label{tab:error_by_model}
\begin{tabular}{lcccccc|c|c}
\toprule
\textbf{Model} & \textbf{Scope} & \textbf{Crit.} & \textbf{Intent} & \textbf{Page} & \textbf{Dep.} & \textbf{Final} & \textbf{Fail} & \textbf{Corr.} \\
 & \textbf{Drift} & \textbf{Mis.} & \textbf{Rew.} & \textbf{Mis.} & \textbf{Err.} & \textbf{Ans.} & & \\
\midrule
Seed-2.0     & 15 & 1 & 2 & 1 & 1 & 0 & 20 & 2 \\
Gemini 3.1   & 11 & 6 & 0 & 0 & 0 & 2 & 19 & 3 \\
Kimi K2.5    &  9 & 4 & 1 & 3 & 5 & 0 & 22 & 0 \\
Qwen3.6+     &  7 & 4 & 2 & 2 & 0 & 5 & 20 & 2 \\
GPT-5.5      &  5 & 6 & 2 & 2 & 1 & 3 & 19 & 3 \\
GLM-5.1      &  5 & 8 & 1 & 1 & 1 & 3 & 19 & 3 \\
DS V4 Pro    &  4 & 8 & 3 & 1 & 2 & 1 & 19 & 3 \\
Claude 4.7   &  2 & 6 & 5 & 2 & 1 & 2 & 18 & 4 \\
\midrule
\textbf{Total} & \textbf{58} & \textbf{43} & \textbf{16} & \textbf{12} & \textbf{11} & \textbf{16} & \textbf{156} & \textbf{20} \\
\bottomrule
\end{tabular}
\end{table}

Three qualitative clusters emerge. \textit{Drift-dominant} models (Seed-2.0 at 75\%, Gemini~3.1 at 58\%) fail primarily in establishing the correct initial retrieval workspace. \textit{Mismatch-dominant} models (DS~V4~Pro and GLM-5.1, each 42\%) reach relevant resources but apply incorrect field definitions. \textit{Mixed-profile} models (GPT-5.5, Claude~4.7) distribute failures more evenly, with Claude~4.7 uniquely exhibiting high intent rewriting (28\%), reflecting its tendency to delegate subtasks to subagents that rephrase the original query. Kimi~K2.5 is the only model with zero correct trajectories across all 22 tasks, and has the highest retrieval-dependency failure rate (23\%), suggesting difficulty maintaining evidence chains across dependent retrieval steps.

\section{Error Profile by Source Family}
\label{app:error_family}

\begin{figure}[h]
\centering
\includegraphics[width=\columnwidth]{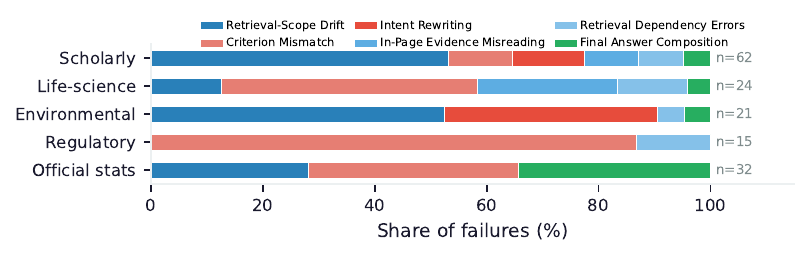}
\caption{Error type distribution per source family. Scholarly and environmental tasks are drift-dominated; regulatory and life-science tasks are mismatch-dominated; official statistics split between criterion mismatch and final answer composition.}
\label{fig:error_by_family}
\end{figure}

Table~\ref{tab:error_by_family} and Figure~\ref{fig:error_by_family} break down error types by source family for the audited trajectory subset. The 22 audited task slots cover five source-family groupings; vulnerability-database tasks are part of the full benchmark but are not included in this trajectory audit.

\begin{table}[h]
\centering
\small
\caption{Error type counts per source-family grouping across 156 failed trajectories in the audited subset.}
\label{tab:error_by_family}
\begin{tabular}{lcccccc|c}
\toprule
\textbf{Source Family} & \textbf{Scope} & \textbf{Crit.} & \textbf{Intent} & \textbf{Page} & \textbf{Dep.} & \textbf{Final} & \textbf{Total} \\
 & \textbf{Drift} & \textbf{Mis.} & \textbf{Rew.} & \textbf{Mis.} & \textbf{Err.} & \textbf{Ans.} & \\
\midrule
Scholarly      & 35 &  7 & 8 & 6 & 5 & 3 & 64 \\
Official stats &  9 & 12 & 0 & 0 & 0 & 11 & 32 \\
Life-science   &  3 & 11 & 0 & 6 & 3 & 1 & 24 \\
Environmental  & 11 &  0 & 8 & 0 & 1 & 1 & 21 \\
Regulatory     &  0 & 13 & 0 & 0 & 2 & 0 & 15 \\
\bottomrule
\end{tabular}
\end{table}

Scholarly archive tasks concentrate failures in retrieval-scope drift (55\%), reflecting the challenge of establishing correct candidate spaces via advanced search filters. Regulatory tasks (CFPB) show 87\% criterion mismatch: agents reach the correct API or data source but apply incorrect query parameters. Official-statistics tasks (Census) split between criterion mismatch (38\%) and final answer composition (34\%), the latter reflecting that Census tasks require assembling multiple API fields into correctly formatted output rows, a step that agents complete semantically but fail to format correctly. Environmental tasks (Water Quality Portal) uniquely combine scope drift (52\%) with intent rewriting (38\%), as agents often simplify multi-HUC comparison tasks into single-station lookups.

\section{Error Type and Task Performance}
\label{app:error_metrics}

\begin{figure}[h]
\centering
\includegraphics[width=\columnwidth]{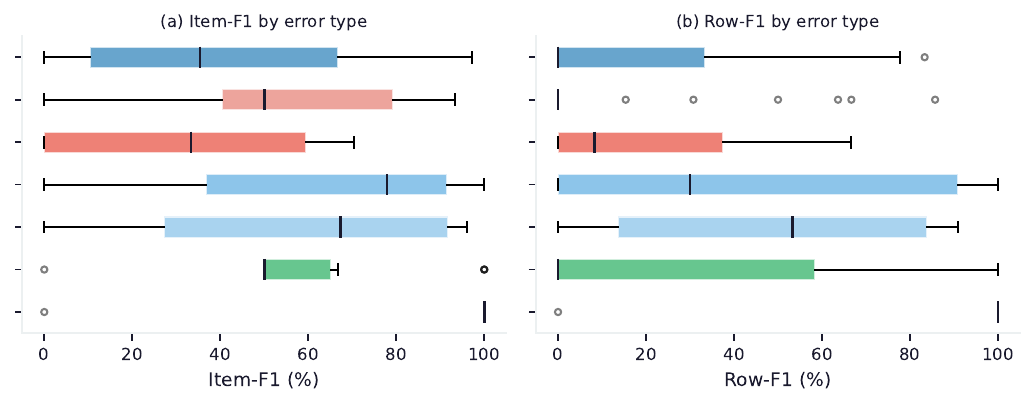}
\caption{Item-F1 and Row-F1 distributions grouped by error type. Criterion mismatch shows the widest Item-F1 vs.\ Row-F1 gap: agents recover many field values but almost no complete rows. Upstream errors (scope drift, intent rewriting) depress both metrics.}
\label{fig:error_vs_metrics}
\end{figure}

Figure~\ref{fig:error_vs_metrics} shows how each error type manifests in Item-F1 and Row-F1 scores. Table~\ref{tab:error_metrics_summary} provides the corresponding statistics.

\begin{table}[h]
\centering
\small
\caption{Mean performance by error type. Upstream errors (scope drift, intent rewriting) produce lower Item-F1, while criterion mismatch preserves moderate Item-F1 but collapses Row-F1.}
\label{tab:error_metrics_summary}
\begin{tabular}{lcccc}
\toprule
\textbf{Error Type} & \textbf{$n$} & \textbf{Item-F1} & \textbf{Row-F1} & \textbf{Gap} \\
\midrule
Retrieval-Scope Drift & 58 & 38.7 & 19.2 & 19.5 \\
Criterion Mismatch    & 43 & 52.5 &  7.3 & 45.2 \\
Intent Rewriting      & 16 & 31.4 & 22.3 &  9.1 \\
In-Page Evidence Misreading    & 12 & 62.8 & 43.1 & 19.7 \\
Retrieval Dependency Errors  & 11 & 58.5 & 49.6 &  8.9 \\
Final Answer Composition  & 16 & 58.5 & 26.4 & 32.1 \\
\midrule
Correct               & 20 & 95.0 & 95.0 &  0.0 \\
\bottomrule
\end{tabular}
\end{table}

The most revealing pattern is criterion mismatch: it achieves moderate Item-F1 (52.5\%) but the lowest Row-F1 among all error types (7.3\%), yielding a 45.2-point gap. This means agents with criterion mismatch errors still extract many locally correct field values, but because the underlying field definitions or decision rules are wrong, almost no row is fully correct. By contrast, retrieval-scope drift depresses both metrics roughly equally (38.7\% Item-F1, 19.2\% Row-F1), consistent with global workspace errors that corrupt all downstream evidence. Correct trajectories confirm the metrics: 95.0\% on both Item-F1 and Row-F1.

\section{Performance vs.\ Task Cardinality}
\label{app:cardinality}

\begin{figure}[h]
\centering
\includegraphics[width=0.65\columnwidth]{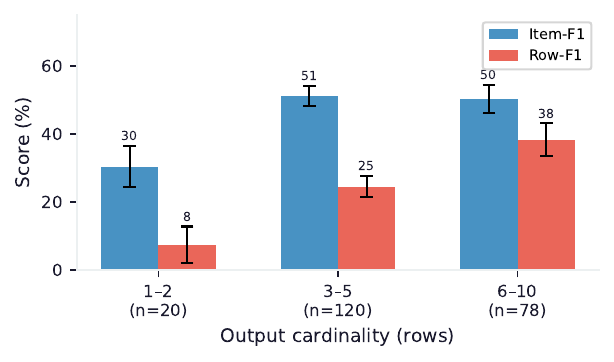}
\caption{Mean Item-F1 and Row-F1 by output cardinality bin across all models. Tasks requiring 1--2 rows are the hardest in the benchmark results, reflecting that low-cardinality cases there tend to involve complex multi-step state operations (e.g., Census, CFPB).}
\label{fig:perf_vs_cardinality}
\end{figure}

Across the 100-task benchmark, expected output cardinality ranges from 2 to 44 rows, with a mean of 6.42 and a median of 4.0. The distribution is concentrated in the low-to-mid range: 26 tasks (26.0\%) require 1--2 rows, 38 (38.0\%) require 3--5 rows, 26 (26.0\%) require 6--10 rows, and 10 (10.0\%) require more than 10 rows; 72 tasks (72.0\%) require at most 7 rows. Figure~\ref{fig:perf_vs_cardinality} groups evaluation instances by the number of rows in the reference answer. Tasks requiring only 1--2 rows are the hardest (Item-F1 = 36.36\%, Row-F1 = 9.38\%). Mid-cardinality bins perform better: 3--5 rows reach 54.43\% Item-F1 and 26.04\% Row-F1, while 6--10 rows reach 52.35\% Item-F1 and 38.77\% Row-F1. Aggregate results are not reported for the 11+ row bin. The hardest low-cardinality cases are concentrated in Census and CFPB, which demand complex multi-step state operations across coupled controls.

% NeurIPS checklist is omitted from the arXiv version.

\end{document}